\documentclass[10pt,twocolumn,letterpaper]{article}

\usepackage{times}
\usepackage{epsfig}
\usepackage{graphicx}
\usepackage{amsmath}
\usepackage{amssymb}
\usepackage{booktabs}

\usepackage[pagebackref=true,breaklinks=true,letterpaper=true,colorlinks,bookmarks=false]{hyperref}

\begin{document}
	
	\title{CMSN: Continuous Multi-stage Network and Variable Margin Cosine Loss for Temporal Action Proposal Generation}
	
	\author{Yushuai Hu\textsuperscript{1}\ \ \ \ \ Yaochu Jin\textsuperscript{1,2}\ \ \ \ \ Runhua Li\textsuperscript{1}\ \ \ \ \ Xiangxiang Zhang\textsuperscript{1}\\
		\textsuperscript{1}Ulucu AI Research \ \ \ \ \ \textsuperscript{2}University of Surrey\\
		{\tt\small \{hyshuai,lrhua,zxxiang\}@ulucu.com, yaochu.jin@surrey.ac.uk}
	}
		
	\maketitle
	
	\begin{abstract}
	Accurately locating the start and end time of an action in untrimmed videos is a challenging task. One of the important reasons is the boundary of an action is not highly distinguishable, and the features around the boundary are difficult to discriminate. To address this problem, we propose a novel framework for temporal action proposal generation, namely Continuous Multi-stage Network (CMSN), which divides a video that contains a complete action instance into six stages, namely Background, Ready, Start, Confirm, End, Follow. To distinguish between Ready and Start, End and Follow more accurately, we propose a novel loss function, Variable Margin Cosine Loss (VMCL), which allows for different margins between different categories. Our experiments on THUMOS14 show that the proposed method for temporal proposal generation performs better than the state-of-the-art methods using the same network architecture and training dataset.
	\end{abstract}
	
	\section{Introduction}
	In recent years, with the development of deep learning, classification accuracy of action recognition on trimmed videos has been significantly improved~\cite{action1,action2,action3-tsn}. But real-world videos are usually untrimmed, making location and action classification in untrimmed videos increasingly important. This task named Action Detection aims to identify the action class, and accurately locate the start and end time of each action in untrimmed videos.
	
	By definition, action detection is similar to object detection. While object detection aims to produce spatial location in a 2D image, action detection aims to produce temporal location in a 1D sequence of frames. Because of the similarity between object and action detection, many methods for action localization are inspired from advances in object detection~\cite{fast,faster}, which first generate temporal proposals and then classify each proposal. However, the performances of these methods remain unsatisfactory. As found in the previous work~\cite{prop_sst,prop_scnn,prop_2_tag,detect_rethinking}, this can mainly be attributed to the different properties between action detection and object detection. For example, the duration of action instances varies a lot, from one second to a few hundred seconds~\cite{detect_rethinking}, and video streaming always contains a lot of information that is not related to the action, making action detection still a challenging problem.

	\begin{figure}[t]
		\centering
		\includegraphics[width=0.49\textwidth]{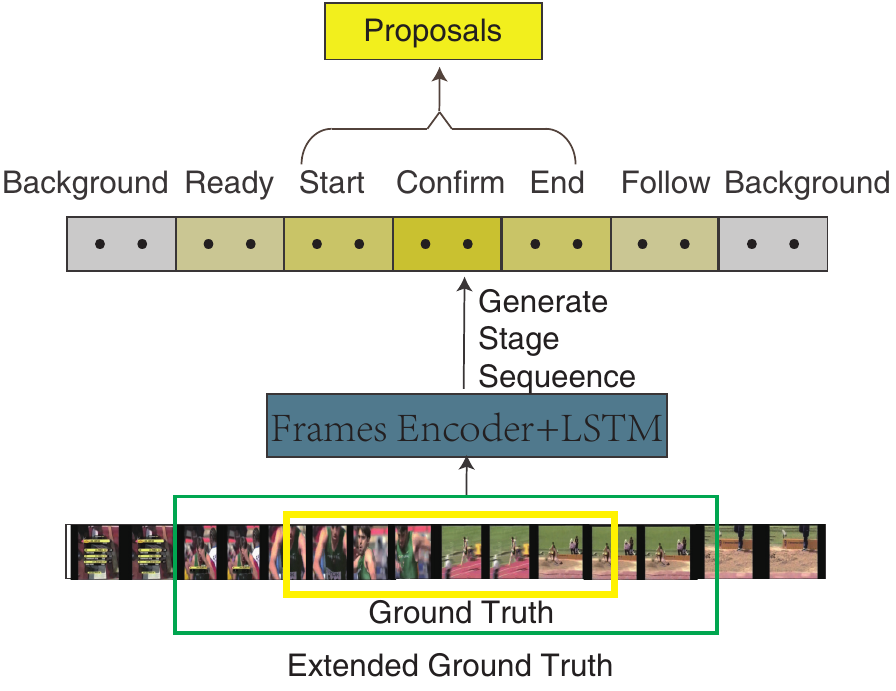}
		\caption{The architecture of our method, firstly we expand the Ground Truth, secondly, the network encodes the frames and generates the action stage categories sequence, finally combines action stages to generate proposals.}
		\vspace{-0.2cm}
		\label{fig:figure1}
	\end{figure}

	
	Because the generation of high-quality proposals is the basis for improving detection performance, in this paper, we mainly focus on the temporal proposal generation. We observe that at the start and end of an action, the action is continuous and does not suddenly start or end. So we believe the probability of the start and end time should not be predicted separately, but should be linked to the front and back to form a continuous sequence of features.

	\begin{figure*}[htb]
		\centering
		\includegraphics[width=0.97\textwidth]{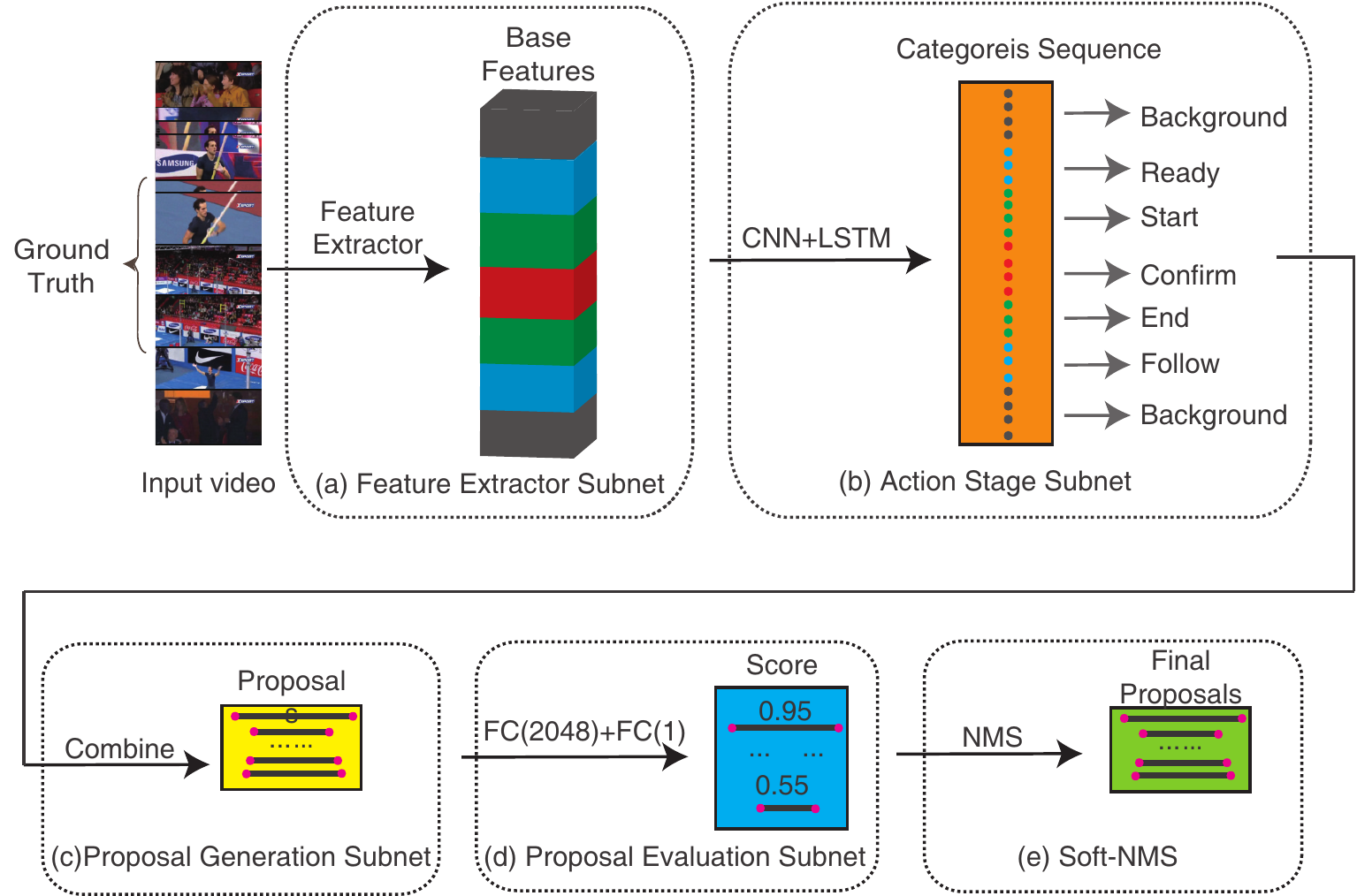}
		\caption{The model architecture. (a) Feature Extractor Subnet (FES) is used for computing feature sequences. (b) The Action Stage Subnet (ASS) takes the features sequence as input, and generates action stage categories sequence corresponding to the input. (c) Proposal Generation Subnet (PGS) combines the action stage sequence to generate proposals and sample features from the features sequence. (d) Proposal Evaluation Subnet (PES) evaluates the IoU of each proposal directly. (e) Soft-NMS}
		\vspace{-0.2cm}
		\label{fig:figure2}
		
	\end{figure*}
	
	We propose a Continuous Multi-stage Network (CMSN), and its architecture is illustrated in Fig.~\ref{fig:figure1}. As shown in the figure, we designed six stage categories based on the video process, and directly predict the category of each frame. 
	The main contributions of our work are two-folds:
	\begin{itemize}
		\item [1)]We introduce a new model, CMSN, which expands a complete action instance and divides it into six stages, namely Background, Ready, Start, Confirm, End, Follow, corresponding to background, a short period before the start, a short period after the start, the middle stage, a short period before the end, a short period after the end, and consider six stages as six categories. Then each action instance could be considered as a continuous category sequence, and the categories around the start and end time are separated. This strategy could improve recall performance and location accuracy, and could handle large variations in action duration to some extent.
		\item [2)]In order to locate the start and end time of an action instance more
		accurately, we propose a new loss function, termed Variable Margin Cosine Loss. The loss function adds a variable angular margin between different classes, which allows similar samples to keep a certain margin and dissimilar samples to have a large margin. 
	\end{itemize}

	\section{Related work}
	Temporal action detection task focuses on predicting action classes, and temporal boundaries of action instances in untrimmed videos. Most action detection methods are inspired by the success of image object detection~\cite{faster,yolo,ssd}. The mainstream methods can be divided into two types, one is a two-stage pipeline~\cite{detect_temporal,detect_r-c3d,detect_rethinking,ssn,detect_temporal2017}, and the other is a single-shot pipeline~\cite{detect_endtoend,detect_single1,detect_single2}. For the two-stage pipeline, the first step is to generate proposals, and the second step is to classify proposals. According to the findings of previous work, under the same conditions, the two-stage method performs better than the single-shot method.

	In a proposal generation task, earlier work~\cite{prop_sliding1,prop_sliding2,prop_sliding3} mainly use sliding windows as candidates. Recently, many methods~\cite{prop_turn,detect_temporal,detect_r-c3d,detect_rethinking} use preset fixed-time anchors to generate proposals. TAG~\cite{ssn} divides an activity instance into three stages, then uses an actionness classifier to evaluate the binary actionness probabilities. SMS~\cite{detect_sms} assumes that each temporal window begins with a single start frame, followed by one or more middle frames, and finally a single end frame. BSN~\cite{prop_bsn} locates temporal boundaries with high probabilities, then evaluate the confidences of candidate proposals generated by these boundaries. BMN~\cite{prop_bmn} proposes the Boundary-Matching mechanism to evaluate confidence scores of densely distributed proposals. MGG~\cite{prop_MGG} proposes a multi-granularity generator to generate temporal action proposals from different granularity perspectives.
	
	Different from the above-mentioned work, we use six stages to represent a complete action instance. Each stage corresponds to a period of an action rather than a frame. We do not just predict the probability of a frame, but predict an action as a continuous multi-stage sequence.	
	

	\section{Proposed Approach \label{approach}}
	We propose a novel network for  generating
	proposals in continuous untrimmed video streams, namely Continuous Multi-stage Network (CMSN). As shown in
	Fig.~\ref{fig:figure2}, CMSN takes a video as input, outputs a sequence of action
	stage categories after Action stage             
	Subnet, and outputs a set of proposals for the input video in the last network.
	Next, we describe Feature Extractor Subnet (FES) and Action Stage Subnet (ASS) in
	Section~\ref{action stage}, Proposal Generation Subnet (PGS) in Section~\ref{proposal}, Proposal Evaluation Subnet (PES) and Soft NMS in Section~\ref{iou}.
	\subsection{Feature Extactor Subnet and Action Stage Subnet \label{action stage}}
	Suppose the input video frames $\mathcal{V} = \{V_1,\ldots,V_t,\ldots,V_n\}$, have dimension
	$[3\times L_i\times H_i\times W_i]$, where $V_t $ denotes the frame at time step $t$, and $n$ is the total number of
	frames in the video. Let $P_i = \{P_s,P_e\}$ denote a proposal, $P_s$ is the start time
	and $P_e$ is the end time, thus the duration of $P_i$ is $d_i = P_e - P_s$. To make full
	use of the context information, we expand the start time $P_s$ to $P_s' = P_s - d_i / 2$, and expand the end time $P_e$ to $P_e' = P_e + d_i/2$, as done in~\cite{ssn}, but we also divide $P_i$ into three segments by $2$ time $(Ps+di/3,Pe-di/3)$: $A=(P_s,P_s+d_i/3)$, $B=(P_s+d_i/3,P_e-d_i/3)$, $C=(P_e-d_i/3,P_e)$. In this way, the expanded proposal is divided into $5$ segments: $\{P_s',P_s\}$, $\{P_s,P_s+d_i/3\}$, $\{P_s+d_i/3,P_e-d_i/3\}$, $\{P_e-d_i/3,P_e\}$, $\{P_e,P_e'\}$, which corresponds to Ready, Start, Confirm, End, Follow, respectively. And with the stage of background, we have six action stage categories: Background, Ready, Start, Confirm, End, Follow. So each frame of the input video corresponds to one of the six action stage categories.	
	

	\textbf{Feature Extractor Subnet }To extract features from a given video, we could use various convolutional networks for action recognition. In our framework, we adapt C3D network~\cite{action_c3d} as FES. For studying the effect of different video features, the two-stream network~\cite{two-stream} will also be experimented. Take the C3D as an example, the input of FES is RGB frames. The input frames $\mathcal{V}$ with dimension $[3\times L_i\times H_i\times W_i]$ are input into FES and output base features. The output features have dimension: $[C_o\times L_o\times H_o\times W_o]$, with the length, width, height being scaled to the original input frames by the network. Suppose the scale of FES on the dimension length is $s_c$, then $L_o = L_i/s_c$.
			
	\textbf{Action Stage Subnet }As shown in Fig.~\ref{fig:figure2}, ASS consists of a CNN network and a LSTM network. The CNN network scales the base features so that they are suited for the LSTM network, containing two convolutional network layers with a kernel size $3$ and hidden size $512$, and one max-pooling layer. The outputs of the CNN network have a size $[L_c\times L_o]$, where $L_c$ is the input channel of the LSTM network. The LSTM network which we used is a bidirectional LSTM~\cite{lstm_bid} having 2 layers, which can make use of the context information to a maximum extent. The outputs of the CNN network feed into the LSTM network, then pass through the classification layer. The final output is an action stage categories sequence. Let $Q= [q_1,q_2,\ldots,q_n]$ be the output sequence, where $n = L_o$. Because the output length $L_o$ is scaled, the frame-level action stage category also needs to be scaled down by the same ratio $s_c$.

	The reason why we divide the proposal into six segments is based on the following observations. Firstly an action never happens suddenly in a continuous video stream. For example, Billiards, before the action starts, the player must move close to the pool table and adjust the position. Secondly, many actions could not be confirmed to happen in a short period after starting, but need to observe for a while. For example, Long Jump, in the run upstage, one could not distinguish whether the action instance is Long-distance or Long jump. Similarly, before an action ends, we often can predict whether an action is coming to an end. That is, the start stage, the middle stage and the end stage of an action often are distinguishable. Thus, we argue that dividing an action into three stages could more effectively describe the action than one stage, and superadding the Ready and Follow stage could more effectively make use of the context information.
		
	We do not just predict the probability of a certain moment, such as the start time, but predict the whole action process and model an action instance as a sequence of action stage categories. In this way, the sequence is continuous as a piece of video, so if an intermediate time instead of the start and end time is detected, we can still estimate the start and end time. The categories around the start and end time are separated, so we can control the margin between them as described in Section~\ref{vmcl}.      
	

	\subsection{Proposal Generation Subnet \label{proposal}}
\begin{figure}[t]
		\centering
		\includegraphics[width=0.48\textwidth]{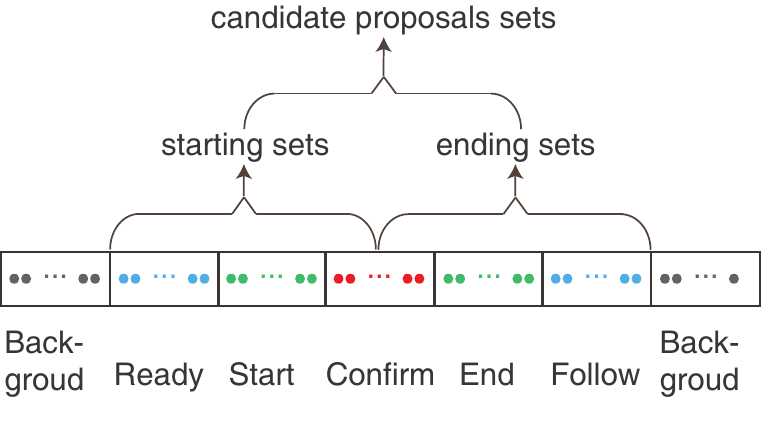}
		\caption{Pipeline of our candidate proposals generation, from action stage categories sequence to candidate proposals sets.}
		\label{fig:prop_generate}
	\end{figure}

	The output of the Action Stage Subnet is an action categories sequence $Q$, which is arranged in the order from Ready to Follow. This corresponds well with the ground truth, but sometimes the order may be erroneous. In the subsequence of the sequence $Q$, let $Q_r= [r_1,r_2,\ldots,r_n]$ denote the Ready sequence, $Q_s= [s_1,s_2,\ldots,s_n]$ denote the Start sequence,  $Q_c= [c_1,c_2,\ldots,c_n]$ denote the Confirm sequence,  $Q_e= [e_1,e_2,\ldots,e_n]$ denote the End sequence, $Q_f= [f_1,f_2,\ldots,f_n]$ denote the Follow sequence. As shown in Fig.~\ref{fig:prop_generate}, we choose start and end locations, and combine them as candidate proposals as follows: 
	\begin{itemize}
		\item [1)]
		We choose a location $t$ as start location, if $t$ belongs to $Q_r$ or $Q_s$ or the first half of $Q_c$. Then we could obtain candidate start point sets $C_{s} = \{r_1,\ldots,s_1,\ldots,c_1,\ldots,c_{n/2}  \}$.
		\item [2)]
		We choose location $t$ as end location, if $t$ belongs to $Q_e$ or $Q_f$ or the second half of $Q_c$. Then we could obtain candidate end point sets $C_{e} = \{c_{n/2},\ldots,c_n,\ldots,e_1,\ldots,f_1,\ldots\}$.
		\item [3)]We combine a start location belonging to $C_{s}$ and an end location belonging to $C_{e}$, if it contains at least one of Start stage and Confirm stage and Follow stage from the start to the end (excluding the Background stage). Finally, we could obtain the candidate proposal point sets $C_{p} = \{(r_1,c_n),\ldots,(r_1,e_1),\ldots,(s_1,f_1),\ldots,(c_1,c_n),\ldots  \}$.
	\end{itemize}

	\subsection{Proposal Evaluation Subnet and Soft NMS \label{iou}}
	The input features of PES is the output of the CNN in ASS with size $[L_c\times L_o\times 1\times 1]$, we intercept the features sequence by the candidate proposal set $P_{s}$, and obtain features sets $P(f)$. Because features in $P(f)$ have different lengths, we use RoI pooling~\cite{detect_r-c3d} to extract the fixed-size volume features. The output of the RoI pooling is fed into two fully connected layers with a hidden size $2048$, and output the score corresponding to set $P(f)$. 
	
	Because the outputs of ASS could correspond well with the ground truth, we set a preset score for each proposal. Let $q_c$ denote a sequence from Ready to Follow, $q_s$ denote the first Start in the sequence,  $q_e$ denote the last End. For a proposal generated from the sequence $q_c$, let $d_s$ denote the distance between the start of the proposal and $q_s$, $d_e$ denote the distance between the end of the proposal and $q_e$. Then we could compute the pre-score as follows:
	\begin{equation}
	pre\verb|-|score = ( 1 - d_s  \times i) \times ( 1 - d_e  \times i) 
	\vspace{-1mm}
	\end{equation}
	where i is the decay rate. Then for each proposal, we use the product of the output score of PES and the pre-score as the final score. Finally, we use non-maximum suppression (NMS)~\cite{nms_soft} to remove redundant proposals.
	\subsection{Training and Prediction \label{train_predict}}
	\textbf{Training }We train the Action Stage Subnet firstly. For an input video $\mathcal{V}$, we need to assign labels to the action stage categories sequence of ASS. We use ground-truth to generate action stage category labels described in Section~\ref{action stage}. Because the ground-truth is at frame level, suppose the scale ratio is $s_c$, as described in Section~\ref{action stage}. We firstly generate frame-level categories sequence, then sample the sequence at $s_c$ fps to generate label sequence corresponding to action stage categories sequence. The used loss function is Variable Margin Cosine Loss, the categories are assigned to $0-6$ corresponding to the stages from Background to Follow. After Action Stage Subnet is trained, we could generate proposals described in Section~\ref{action stage}. Then we select proposals by Intersection-over-Union (IoU) with some ground-truth activity as follows: firstly select all proposals with IoU higher than $0.6$ as Big, suppose the number of Big is $n$; secondly, select proposals with IoU higher than $0.2$ and lower than $0.6$ with the same number $n$ as Middle; finally, select proposals with IoU lower than $0.2$ with the number $n$ as Smaller. Then train the IoU Evaluation Subnet with Smooth L1 loss~\cite{fast}.
	
	\textbf{Prediction }For a video to be predicted, we first sample to generate some frames sequences with the same length in training $L_i$, then we could generate action stage categories sequence with the architecture described in Section~\ref{action stage}. Because a complete proposal may be truncated at the beginning or end of frames sequence, we have half overlap between contiguous sample frames, finally we can generate proposals described in Section~\ref{proposal} and Section~\ref{iou}.

	\section{Variable Margin Cosine Loss}
	\subsection{Softmax Loss and Variations}
	Softmax loss is widely used in classification tasks and can be formulated as follows:	
	\begin{equation}
	L_s=-\frac{1}{m}\sum_{i=1}^{m}\log\frac{e^{W^T_{y_i} x_i+b_{y_i}}}{\sum_{j=1}^{n}e^{W^T_j x_i+b_j}},
	\label{eq:softmax}
	\vspace{-1mm}
	\end{equation}
	where $x_i$ denotes the input feature vector of the $i$-th sample, corresponding to the $y_i$-th class. $W_j\in\mathbb{R}^d$ denotes the $j-$th column of the weight vector $W \in \mathbb{R}^{d \times n}$ and $b_j\in\mathbb{R}^n $ is the bias term. The size of mini-batch and the number of class is $m$ and $n$, respectively. Under binary classification case, the decision boundary of Softmax loss is: $w_1 x+b_1=w_2 x+b_2$. Obviously, the boundary does not have a gap between a positive sample and a negative sample. To solve this problem, some variants were proposed, such as CosFace (LMCL)~\cite{cosface}, ArcFace~\cite{arcface} and A-softmax~\cite{sphereface}. Take LMCL as an example, the loss function is formulated as follows:
	\begin{equation}\label{1}
	L_{lmc} = \frac{1}{N}\sum_{i}{-\log{\frac{e^{s (\cos(\theta_{{y_i}, i}) - m)}}{e^{s (\cos(\theta_{{y_i}, i}) - m)} + \sum_{j \neq y_i}{e^{s \cos(\theta_{j, i})}}}}},
	\vspace{-1mm}
	\end{equation}	
	subject to	
	\begin{equation}
	W = \frac{W^*}{\Vert{W^*}\Vert}, x = \frac{x^*}{\Vert{x^*}\Vert}, cos(\theta_j,i) = {W_j}^Tx_i
	\vspace{-1mm}
	\end{equation}
	where $N$ is the numer of training samples, $s$ is the scale factor, $x_i$ is the $i$-th feature vector corresponding to the ground-truth class of $y_i$,
	the $W_j$ is the weight vector of the $j$-th class, and $\theta_j$ is the angle between $W_j$ and $x_i$. The decision boundary is: $cos(\theta_1) -m = cos(\theta_2)$. 
	
	All these variants have a margin $m$ between positive and negative samples, and could perform well on some classification tasks such as face recognition. But they do not consider the margin value between different negative samples and the same positive sample, that is, the margin is the same for all negative samples and positive samples. There is a similarity relationship between different sample categories, denoted by $S(i,j)$. We define this relationship $S(i,j)$ as the distance between two categories in the sample space, and define the features set for classification as feature space. Then these variations could not maintain the distance $S(i,j)$ when mapped from the sample space to the feature space. We believe that it is unreasonable to use the same margin for positive and negative samples and hypothesize that the margin should vary depending on the distance $S(i,j)$, i.e., the larger $S(i,j)$ is, the larger the margin should be.  
	



	\subsection{Variable Margin Cosine Loss \label{vmcl}}
	To achieve the above goal, we propose the Variable Margin Cosine Loss (VMCL). Formally, let $i$ denote $i$-th class, $j$ denote $j$-th class, we define VMCL as follows:
	
	\begin{equation}\label{eq_1}
	L_{v} = \frac{1}{N}\sum_{i}{-\log{\frac{e^{s (\cos(\theta_{{y_i}, i}))}}{e^{s (\cos(\theta_{{y_i}, i}))} + \sum_{j \neq y_i}{e^{s (\cos(\theta_{j, i}) + f(j,i))}}}}}
	\end{equation}
	where $f(j,i)$ is variable that determines the margin between $i$ and $j$. The decision boundary is: $cos(\theta_i) = cos(\theta_j) + f(j,i)$. $f(j,i)$ could have various forms as long as it guarantees that $f(j,i)$ is bigger when the distance $S(i,j)$ between $i$ and $j$ is bigger. Without loss of generality, we define $f(j,i)$ as follows:
	\begin{equation}\label{eq_distance}
	f(j,i) = \left| v(j)-v(i) \right| \times m + n
	\vspace{-1mm}
	\end{equation}
	where $n$ is the minimum margin, $m$ controls the growth rate of the interval. $v(i)$ is the predefined value of class $i$. The setting for $v(i)$ is based on the distance $S(i,j)$ between two classes.

	\begin{figure}[t]
		\centering
		\includegraphics[width=0.47\textwidth]{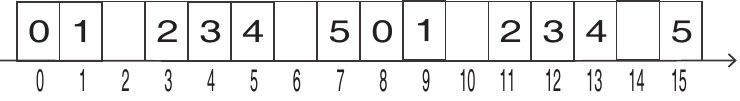}
		\caption{Preset values for different classes, inside the box is the class, and the below is the corresponding value. For example, class $2$ corresponds to value $3,11$.}
		\vspace{-0.2cm}
		\label{fig:fig_margin}
	\end{figure}
	
	When classifying the action stage category, in order to localize the start and end time more accurately, we increase the margin between the ready stage and start stage, end-stage and follow stage. Let $0$, $1$, $2$, $3$, $4$, $5$, $6$ denote Background, Ready, Start, Middle, End, Follow, respectively. We preset $v(i)$ as Fig.~\ref{fig:fig_margin}, each category corresponds to one or more values, then we calculate $f(j,i)$ as follows:
	\begin{equation}
	f(j,i) = \min (\left| v(j)-v(i) \right|) \times m + n  
	\vspace{-1mm}
	\end{equation} 
	This setting makes a large margin between Ready and Start, End and Follow. And the setting also makes a bigger margin when the distance between the two stages is bigger. For example, let $m =0.1$ and $n=0.15$, the margin between Ready and other stages is $[0.15,0.0,0.25,0.35,0.45,0.25]$, which corresponds to stages from Background to Follow. 
	\subsection{Discussions}
	Compared with Softmax Loss and Variations, VMCL considers the different margin values between different categories, and maps the distance $S(i,j)$ in the sample space to the variable margin $f(j,i)$ in the feature space. And because $f(j,i)$ is variable, we could set a large margin even if the distance $S(i,j)$ is small. VMCL can also have other forms. For example, based on ArcFace, VMCL can be defined as follows:
	\begin{equation}\label{eq_arcface}
	L_{v}=\frac{1}{N}\sum_{i}{-\log{\frac{e^{s (\cos(\theta_{{y_i}, i}))}}{e^{s (\cos(\theta_{{y_i}, i}))} + \sum_{j \neq y_i}{e^{s (\cos(\theta_{j, i} - f(j,i)) )}}}}}
	\end{equation}
	the symbols are the same as Equation~\ref{eq_1}.
	\begin{figure}[t]
		\centering
		\includegraphics[width=0.49\textwidth]{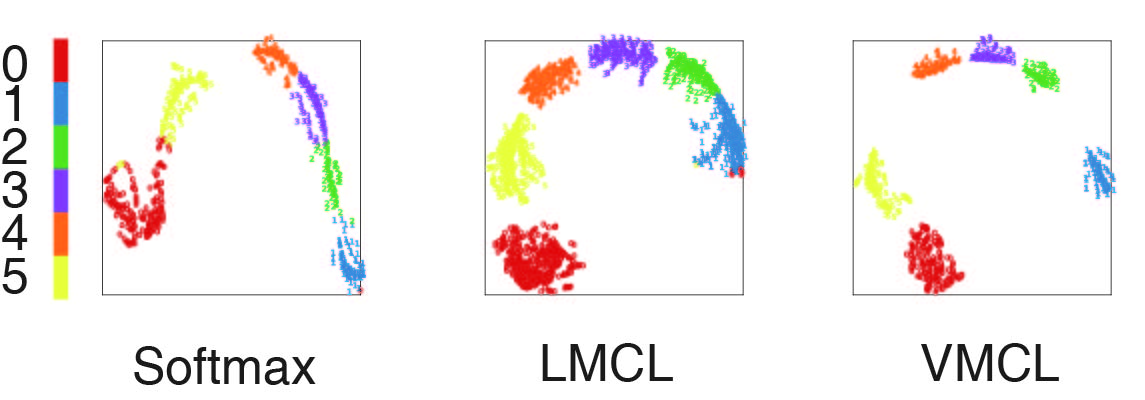}
		\caption{Visualization of features learned with different loss functions. We project the features used in the ASS onto the angular space. $0$, $1$, $2$, $3$, $4$, $5$, $6$ denote Background, Ready, Middle, End, Follow, respectively. }
		\vspace{-0.2cm}
		\label{fig:fig_loss}
	\end{figure}
	To better understand the impact of VMCL on classification, we did a pilot study on different loss functions. We take out the features before the classification layer in the ASS and reduce the dimension to the 2-dimensional angular space. As shown in Fig.~\ref{fig:fig_loss}, Softmax does not have a significant margin between the different categories. LMCL has a large margin and the margin is the same. VMCL has a varying margin between different categories. In particular, according to our settings, there is a larger margin between Ready, Follow and Start, End. This distribution makes the features around the start and end time more distinguishable, and theoretically can improve the positioning accuracy. 
	\section{Experiments}
	\subsection{Dataset and Experimental Settings}
	\textbf{Dataset }We compare our method with the state-of-the-art methods on the temporal action detection benchmark of THUMOS14 ~\cite{data_thumos}. THUMOS14 dataset includes 200 and 213 temporal annotated untrimmed videos with 20 action classes in validation and test sets, respectively. On average, each video contains more than 15 action instances. Because the training set of THUMOS14 contains trimmed videos, so following the settings in previous work, we use 200 untrimmed videos in the validation set to train our model and evaluate on the test set.
	
	\textbf{Evaluation metrics }For the temporal action proposal generation task, Average Recall (AR) calculated with multiple IoU thresholds is typically used as the evaluation metrics. In our experiments, we use the IoU thresholds set $[0.5:0.05:1.0]$ on THUMOS14. And we also evaluate AR with Average Number of proposals (AN) on THUMOS14, which is denoted as AR@AN.
	
	\textbf{Implementation details }We experiment with two feature extraction methods: C3D network~\cite{action_c3d} and two-stream network~\cite{two-stream}. For the C3D network, our implementation uses RGB frames continuously extracted from a video as input. The length of the input is set to $768$, all video frames are resized to $128 \times 171$, then cropped to $112 \times 112$. Since the GPU memory is limited, the mini-batch size is set to $1$. So the input dimensions are $1 \times 3 \times 768 \times 112 \times 112$. The FES adopts the convolutional layers (conv1a to conv5b) of C3D~\cite{action_c3d} pre-trained on ActivityNet-1.3~\cite{data_activitynet} training set, and freeze the first two convolutional layers. The learning rate is set to $0.0001$ and the weight decay parameter is $0.0005$, the dropout rate is $0.5$. The prescore decay rate is $0.09$, n is $0.15$, m is $0.1$. We first train the Action Stage Subnet for three epochs, then train Iou Evaluation Subnet for four epochs while freezing the Feature Extractor Subnet and the Action Stage Subnet.
	
		\begin{table}[tbp]
		\setlength{\abovecaptionskip}{0.1cm}
		\setlength{\belowcaptionskip}{0.3cm}
		
		\centering
		\caption{Comparisons on THUMOS14 in terms of AR@AN using the C3D network.}
		\setlength{\tabcolsep}{1.25mm}{
		\begin{tabular}{ccccc}
			\toprule
			Network & Method  		& @50 & @100  & @200     \\
			\hline 
			C3D & DAPs \cite{prop_daps} 		& 13.56	& 23.83 &  33.96   \\
			C3D & SCNN-prop \cite{prop_scnn} 	& 17.22 & 26.17 &  37.01  \\
			C3D & SST \cite{prop_sst}			& 19.90 & 28.36  &  37.90   \\
			C3D & TURN \cite{prop_turn} 			& 19.63 & 27.96 &  38.34  \\
			C3D & MGG\cite{prop_MGG} 	& 29.11 & 36.31 &  44.32   \\
			C3D & BSN+Greedy-NMS\cite{prop_bsn} 	& 27.19 & 35.38 &  43.61   \\
			C3D & BSN+Soft-NMS\cite{prop_bsn}	& 29.58 & 37.38 & 45.55    \\
			C3D & BMN+Greedy-NMS\cite{prop_bmn} 	& 29.04 & 37.72 & 46.79    \\
			C3D & BMN+Soft-NMS\cite{prop_bmn}	& 32.73 & 40.68 & 47.86     \\		
			\hline 
			C3D & CMSN+Greedy-NMS 	&  {\bf40.45} & 46.48 &  52.23   \\
			C3D & CMSN+Soft-NMS		& 40.40 & {\bf 46.71} &  {\bf 52.46}    \\ 
			\bottomrule
		\end{tabular}}
		\label{table_comparison_1}
		\vspace{0.1cm}
	\end{table}
	\begin{table}[tbp]
		\setlength{\abovecaptionskip}{0.1cm}
		\setlength{\belowcaptionskip}{0.3cm}
		\centering
		\caption{Comparisons on THUMOS14 in terms of AR@AN using the 2-Stream network.}
		\setlength{\tabcolsep}{1.25mm}{
		\begin{tabular}{ccccc}
			\toprule
			Network & Method  		& @50 & @100  & @200     \\
			\hline 
			Flow & TURN \cite{prop_turn} 		& 21.86 & 31.89 & 43.02   \\ 
			2-Stream & TAG \cite{prop_2_tag} 		& 18.55 & 29.00  &  39.61 \\
			2-Stream & CTAP  \cite{prop_2_ctap} 		& 32.49 & 42.61  &  51.97 \\
			2-Stream & MGG  \cite{prop_MGG} 		& 39.93 &47.75  &  51.97 \\
			2-Stream & BSN+Greedy-NMS\cite{prop_bsn} & 	 35.41 & 43.55 &  52.23  \\
			2-Stream & BSN+Soft-NMS\cite{prop_bsn} &  37.46 &  46.06 &  53.21   \\
			2-Stream & BMN+Greedy-NMS\cite{prop_bmn} & 37.15 & 46.75 & 54.84  \\
			2-Stream & BMN+Soft-NMS\cite{prop_bmn} &  39.36 & 47.72 & 54.70   \\
			\hline 
			2-Stream & CMSN+Greedy-NMS 	& 43.24 & 50.22 &  56.11   \\
			2-Stream & CMSN+Soft-NMS	& {\bf 43.77} & {\bf 50.55} &  {\bf 56.48}    \\ 
			\bottomrule
		\end{tabular}}
		\label{table_comparison_2_stream}
		\vspace{0.1cm}
	\end{table}
	
	\begin{table}[tbp]
		\setlength{\abovecaptionskip}{0.1cm}
		\setlength{\belowcaptionskip}{0.1cm}
		\centering
		\caption{Comparisons of different pre-trained C3D model on THUMOS14 in terms of AR@AN. }
		\begin{tabular}{cccc}
			\toprule
			pre-trained model  		& @50 & @100  & @200     \\
			\hline 
			UCF-101 	& 39.87 & 46.08 &  52.07   \\
			Activitynet	& 40.40 & 46.71 &   52.46    \\ 
			\bottomrule
		\end{tabular}
		\label{table_comparison_2}
		\vspace{0.1cm}
	\end{table}
	
	\begin{table}[tbp]
		\setlength{\abovecaptionskip}{-0.05cm}
		\setlength{\belowcaptionskip}{0.1cm}
		\centering
		\caption{Comparison with and without PES on THUMOS14 in terms of AR@AN.}
		\begin{tabular}{ccccc}
			\toprule
			&Network & @50 & @100  & @200     \\
			\hline 
			Without PES &C3D	& 38.52 & 42.41 &  49.35   \\
			With PES &C3D	& 40.40 & 46.71  & 52.46    \\ 
			\hline 
			Without PES &2-Stream	& 39.04 & 47.13 &  53.53   \\
			With PES &2-Stream	&  43.77 & 50.55 &  56.48    \\
			\bottomrule
		\end{tabular}
		\label{table_comparison_3}
		\vspace{0.1cm}
	\end{table}

	\begin{table}[tbp]
	\setlength{\abovecaptionskip}{-0.05cm}
	\setlength{\belowcaptionskip}{0.1cm}
	\centering
	\caption{Comparisons of different IoU thresholds on THUMOS14 in terms of Recall@AN=100 when using the 2-Stream network.}
	\begin{tabular}{ccccc}
		\toprule
		ASS setting  & IoU=0.70  & IoU=0.74 & IoU=0.8 \\
		\hline 
		CNN-LSTM  & 50.23 &  50.47  &49.57 \\
		AvgPool-LSTM  & 50.38  & 50.55  & 49.69 \\ 

		\bottomrule
	\end{tabular}
	\label{table_comparison_iou}
	\vspace{-0.2cm}
	\end{table}
	
	
	\begin{figure}[t]
		\begin{center}
			\includegraphics[width=0.8\linewidth]{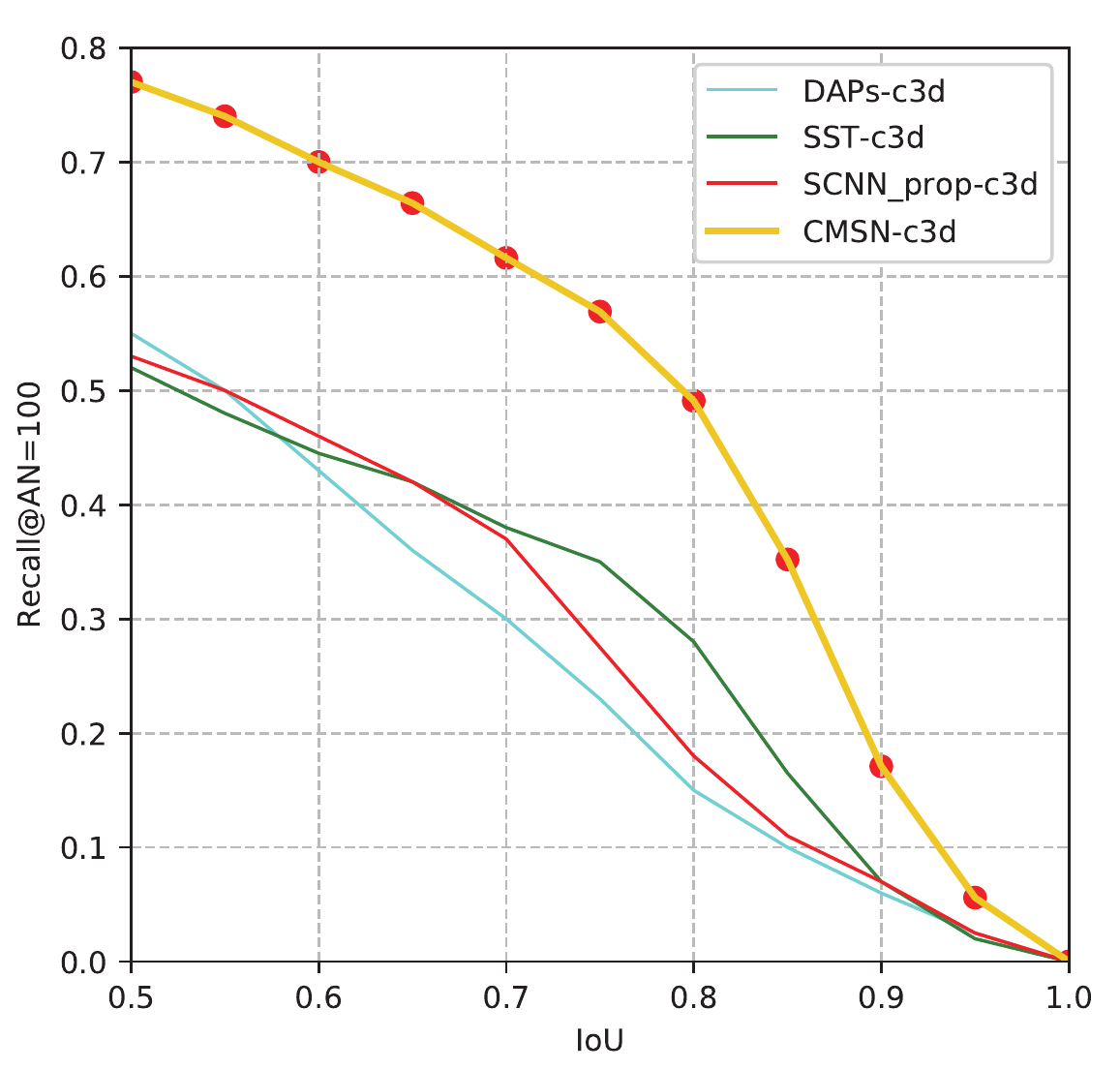}
		\end{center}
		\caption{Recall@AN=100 curves of different methods using C3D network on THUMOS14.}
		\label{fig:recall_100_c3d}
		\vspace{-0.4cm}		
	\end{figure}
	\begin{figure}[t]
		\begin{center}
			\includegraphics[width=0.78\linewidth]{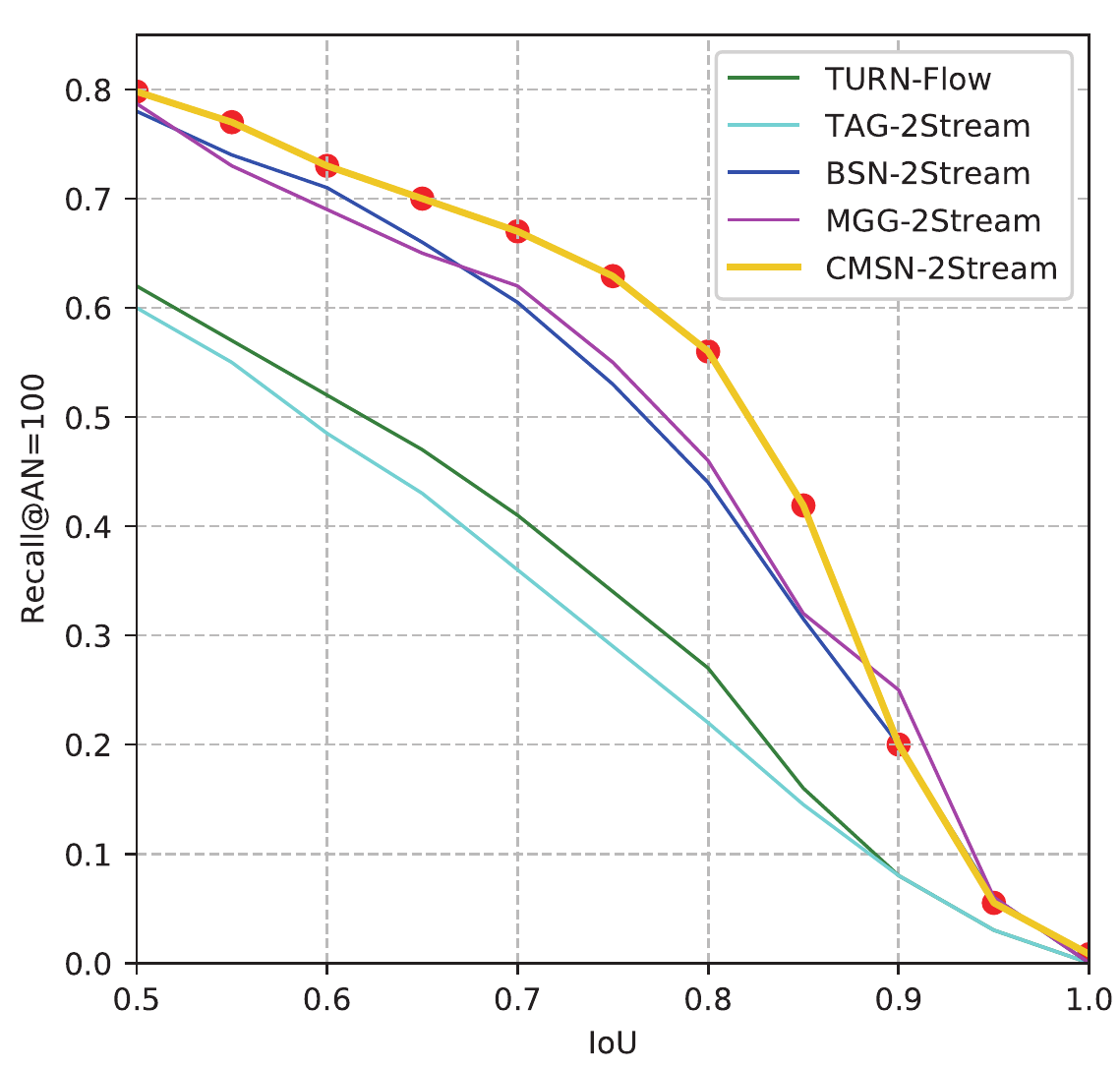}
		\end{center}
		\caption{Recall@AN=100 curves of different methods using 2-Stream network on THUMOS14.}
		\label{fig:recall_100_2stream}
		\vspace{-0.1cm}
	\end{figure}
	\begin{figure}[t]
		\begin{center}
			\includegraphics[width=0.8\linewidth]{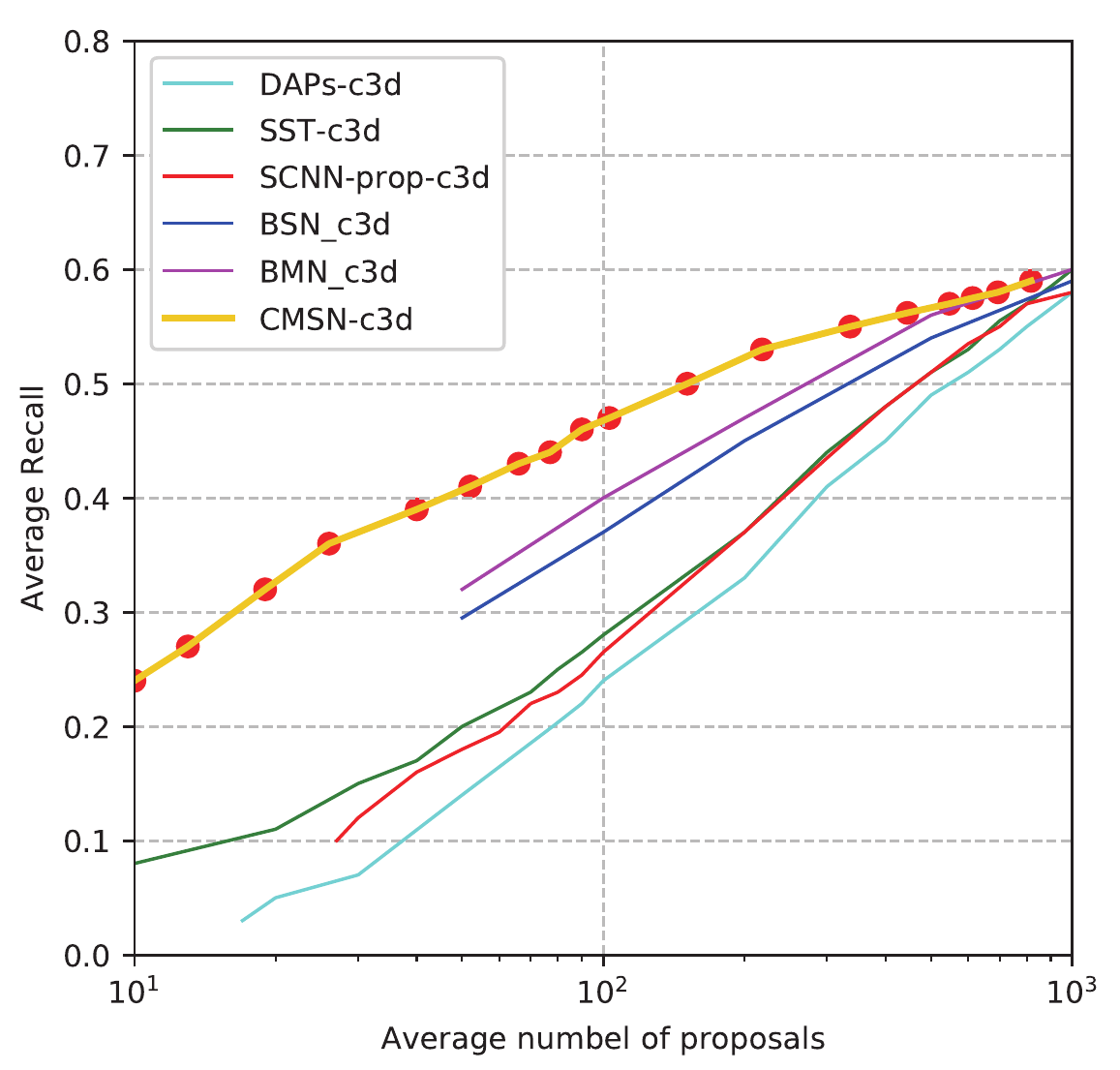}
		\end{center}
		\caption{AR-AN curves of different methods using C3D network on THUMOS14.}
		\label{fig:recall_all_c3d}
		\vspace{-0.4cm}
	\end{figure}
	\begin{figure}[t]
		\begin{center}
			\includegraphics[width=0.8\linewidth]{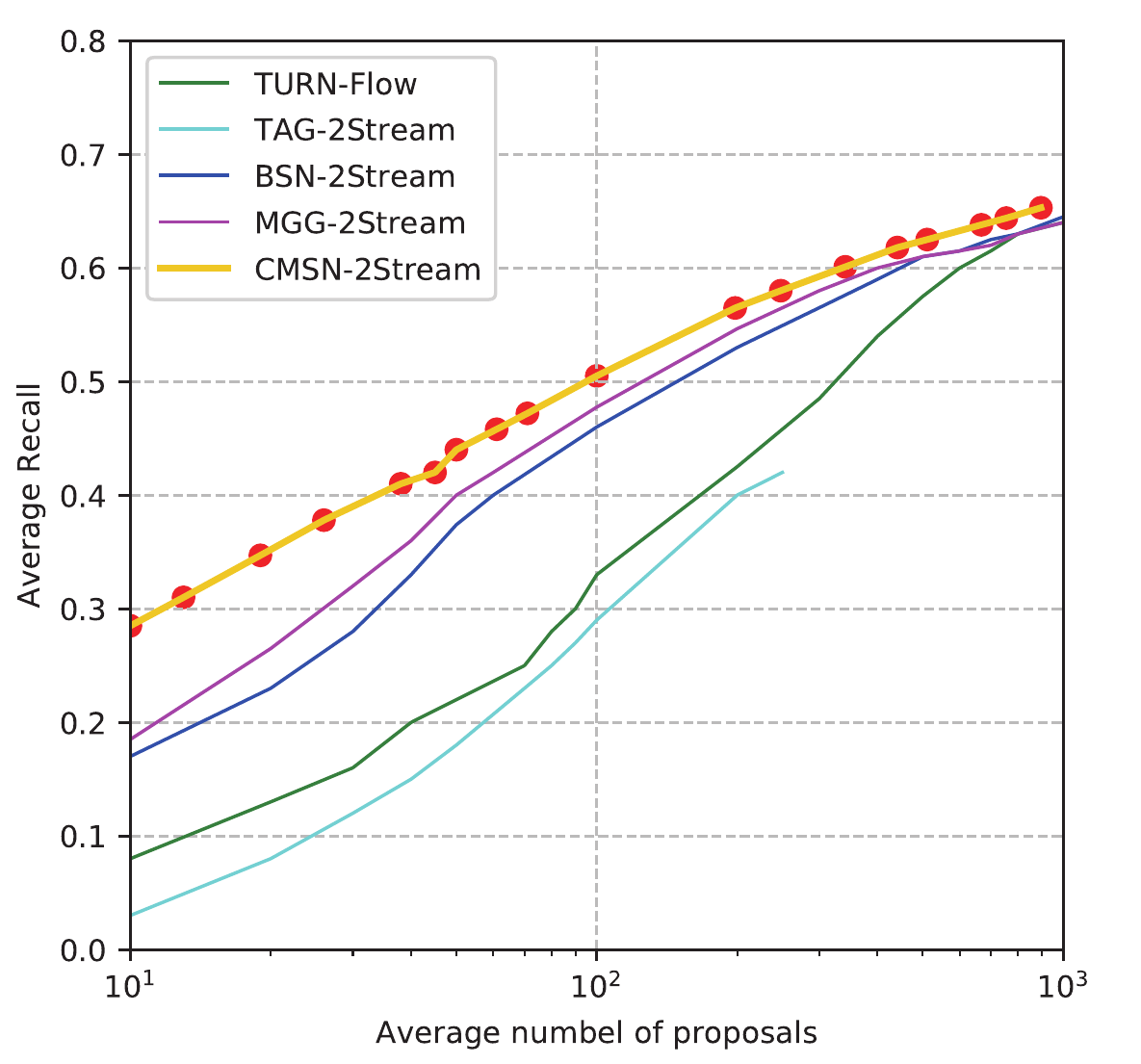}
		\end{center}
		\caption{AR-AN curves of different methods using 2-Stream network on THUMOS14.}
		\label{fig:recall_all_2stream}
		\vspace{-0.4cm}
	\end{figure}
	
	

	For the 2-Stream network, we use the architecture described in~\cite{two-stream_xiong}. The network uses BN-Inception as the temporal network, and uses ResNet as the spatial network. We use the network pre-trained on ActivityNet-1.3~\cite{data_activitynet}. The extracted features are concatenated from the output of the last fully-connected layer, with a dimension of $400$. Because the shape of the extract features has changed, we fine-tune the structure of the ASS. We change the CNN in the ASS from two convolutional layers and one max-pooling layer to one average-pooling layer. The rest of the network structure remains unchanged. We first train the Action Stage Subnet for sixteen epochs, then train Iou Evaluation Subnet for sixteen epochs. The other settings are the same as for the C3D network.

	\subsection{Comparisons}
	\textbf{Comparison of AR@AN } We first evaluate the average recall performance with numbers of proposals (AR@AN) and area under the AR-AN curve. Table~\ref{table_comparison_1} summarizes all comparative results obtained by C3D network on the test set of THUMOS14. It is observed that our method outperforms other methods when AN ranges from 50 to 200. For example, for AR@100, our method improves the performance from the previous record 40.68\% to 46.71\%. Specifically, for AR@50, our method significantly improves the performance from the previous record 32.73\% to 40.45\%, and surpasses the previous results which use the 2-Stream network. Table~\ref{table_comparison_2_stream} shows the results using 2-Stream network. It can also be seen that CMSN has the best performance. These results suggest that our method is efficient and does not rely on a specific feature extraction network. 
	
	\textbf{Comparison of different NMS }Because most previous work adopted Greedy-NMS~\cite{nms_greed} for redundant proposals suppression, we also experiment with the effect of different NMS methods on the results. Results in Table~\ref{table_comparison_1} and Table~\ref{table_comparison_2_stream} show that the results of different NMS methods are comparable, suggesting that our method does not rely on a specific NMS method.
	

	\textbf{Comparison of AR@AN cureves }
	Fig.~\ref{fig:recall_100_c3d} and Fig.~\ref{fig:recall_100_2stream} illustrate the Recall@AN=100 curves of different methods on THUMOS14. When using the C3D networks, the performance of our method is significantly better than that of the compared ones. When using the 2-stream network, our method also achieved the best performance. Fig.~\ref{fig:recall_all_c3d} and Fig.~\ref{fig:recall_all_2stream} illustrate the average recall against the average number of proposals curve of THUMOS14. It can be seen that our methods have achieved the best performance. Especially in the low AN region, our method shows clear advantages. These results suggest that our method could generate proposals of higher quality. 

	
	\textbf{Comparison of different pre-trained models}
	To more accurately evaluate the effects of our model and the proposed loss function, we present experimental results of CMSN with different pre-trained C3D model in Table~\ref{table_comparison_2}. In Table~\ref{table_comparison_2}, UCF-101 denotes the model is pre-trained on UCF-101~\cite{dataset_ucf101}, Activitynet denotes the model is pre-trained on ActivityNet-1.3 training set. For the C3D model pre-trained on UCF-101, we freeze all convolutional layers (conv1a to conv5b), the other settings are the same as when using the C3D model pre-trained on ActivityNet-1.3. As shown in Table~\ref{table_comparison_2}, the results of different pre-trained C3D models are very close, suggesting that our method does not depend on a specific pre-trained model.


	\begin{table}[tbp]
		\setlength{\abovecaptionskip}{-0.05cm}
		\setlength{\belowcaptionskip}{0.1cm}
		\centering
		\caption{Comparison of the effect of VMCL on THUMOS14 in terms of AR@AN.}
		\begin{tabular}{cccc}
			\toprule
			& @50 & @100  & @200     \\
			\hline 

			ASS with Softmax Loss	& 34.73 & 38.89 &  42.58   \\
			ASS with LMCL~\cite{cosface} 	& 34.82 & 38.99 &  42.25   \\
			ASS with VMCL	& 38.52 & 42.41 &  49.35    \\ 
			\bottomrule
		\end{tabular}
		\label{table_comparison_vmcl}
		\vspace{0.1cm}
	\end{table}

	\begin{table}[tbp]
	\setlength{\abovecaptionskip}{-0.05cm}
	\setlength{\belowcaptionskip}{0.1cm}
	\centering
	\caption{Comparison different settings of VMCL on THUMOS14 in terms of AR@AN.}
	\begin{tabular}{ccccc}
		\toprule
		n &m & @50 & @100  & @200     \\
		\hline 
		
		0.1 &0.1	& 37.61 & 41.85 &  47.81  \\
		0.15 &0.1 	&{\bf 38.52}  & 42.41 &  {\bf49.35}   \\
		0.18 &0.12	&  37.83 &  {\bf42.59} &  48.60    \\ 
		0.25 &0.8	&  37.79 &  42.30 &  47.59    \\
		\bottomrule
	\end{tabular}
	\label{table_comparison_mn}
	\vspace{-0.4cm}
	\end{table}
	
	\textbf{Comparison with or without PES }
	Table~\ref{table_comparison_3} shows the results of CMSN with and without PES, where without PES means that the Proposal Evaluation Subnet and the NMS Subnet are removed. The results without PES are close to adding PES and NMS Subnet. These results suggest that the performance enhancement can mainly be attributed to ASS and VMCL.
	
	\textbf{Comparison different IoU thresholds }
	Table~\ref{table_comparison_iou} shows the results of different IoU thresholds and different settings. We use the 2-Stream network as the Feature Extractor Subnet. CNN-LSTM means that the CNN of the ASS contains one convolutional layer, Avgpool-LSTM means that the CNN of the ASS contains one avg-pool layer. It can be seen that these results are comparable and both exceed the previous works. These results suggest that our method is robust to configuration of the network. 
	

	\subsection{Ablative Study for VMCL}
	To demonstrate the effectiveness of different settings in VMCL, we run several ablations to analyze VMCL. All experiments in this ablation study are performed on the THUMOS14 dataset with the C3D network. And for removing the effect of the Proposal Evaluation Subnet, we only compare the results after removing PES and the Soft-NMS Subnet.
	
	\textbf{Effect of VMCL }Table~\ref{table_comparison_vmcl} shows the results of different loss functions. With Softmax Loss, we still get a good result, which suggests that the architecture of ASS is reasonable. With LMCL, we have the same margin between different classes, and the performance of LMCL is worse than VMCL. VMCL performs the best and has significant advantages over other loss functions. These results suggest that ASS is effective and VMCL could achieve better performance.
	                           
	\textbf{Effect of m and n }Since the parameters $m$ and $n$ of VMCL control the decision boundary, we conduct experiments with different values. As shown in Table~\ref{table_comparison_mn}, the preformance is better when $n = 0.15, m = 0.1$ and  $n = 0.18, m = 0.12$, and performance degrades when $n = 0.1, m = 0.1$ and $n = 0.25, m = 0.8$. This means that a too small or too large margin will result in performance degradation. 
	
	VMCL can also be used for other tasks. For example, when the target probability distribution is in the interval $[0-1]$, VMCL can be used to replace the Mean Squared Error. We can consider the probability in the $[0-1]$ interval as 100 categories, then we can use Equation~\ref{eq_1} and Equation~\ref{eq_distance} to calculate the classification loss.
	
	
	
	\section{Conclusions}
	In this paper, we proposed a Continuous Multi-stage Network (CMSN) and a Variable Margin Cosine Loss (VMCL) for temporal action proposal generation. The CMSN directly predicts the entire sequence of action processes rather than predicting the probability of a frame. The VMCL could arbitrarily adjust the distance between different classes in the feature space. The results on large-scale benchmarks THUMOS14 suggest that our proposed method could significantly improve the location accuracy.

	{\small
		\bibliographystyle{ieee_fullname}
		\bibliography{cmsn}
	}
\end{document}